\documentclass[letterpaper, 10 pt, conference]{ieeeconf}  % Comment this line out if you need a4paper

\IEEEoverridecommandlockouts                              % This command is only needed if 
\usepackage{amsmath}
\usepackage{amssymb}
\usepackage{tabularx,booktabs}
\newcolumntype{C}{>{\centering\arraybackslash}X} % centered version of "X" type
\newcolumntype{S}{>{\centering\arraybackslash\hsize=.5\hsize}X} % centered and small version of "X" type
\newcolumntype{s}{>{\hsize=.5\hsize}X} % small version of "X" type

\usepackage{fixltx2e}
\usepackage{amsmath}
\usepackage{amssymb}

\usepackage{subfig}
\usepackage{float}
\usepackage{multirow}
\usepackage{hyperref}
\usepackage{color}
\usepackage{tikz}
\usepackage{graphicx}
\usepackage{graphics}
\usetikzlibrary{plotmarks}
\usepackage{pgfplots}

\let\labelindent\relax
\usepackage[inline]{enumitem}

\usepackage{cite}

\usepackage{gensymb}

\title{\LARGE \bf DynaSLAM II: Tightly-Coupled Multi-Object Tracking and SLAM}

\author{Berta Bescos, Carlos Campos, Juan D. Tard\'{o}s and Jos\'{e} Neira% <-this % stops a space
\thanks{This work has been supported by the Spanish Ministry of Economy and Competitiveness (projects PID2019-108398GB-I00, PGC2018-096367-B-I00 and FPI grant BES-2016-077836).}%
\thanks{All authors are with the Instituto de Investigaci\'on en Ingenier\'ia de Arag\'on (I3A), Universidad de Zaragoza, Zaragoza 50018, Spain {\tt\small \{bbescos,campos,tardos,jneira\}@unizar.es}}%
}

\begin{document}

\maketitle
\thispagestyle{empty}
\pagestyle{empty}

%%%%%%%%%%%%%%%%%%%%%%%%%%%%%%%%%%%%%%%%%%%%%%%%%%%%%%%%%%%%%%%%%%%%%%%%%%%%%%%%
\begin{abstract}
The assumption of scene rigidity is common in visual SLAM algorithms. 
However, it limits their applicability in populated real-world environments. 
Furthermore, most scenarios including autonomous driving, multi-robot collaboration and augmented/virtual reality, require explicit motion information of the surroundings to help with decision making and scene understanding. 
We present in this  paper DynaSLAM~II, a visual SLAM system for stereo and RGB-D configurations that tightly integrates the multi-object tracking capability.

DynaSLAM~II makes use of instance semantic segmentation and of ORB features to track dynamic objects. 
The structure of the static scene and of the dynamic objects is optimized jointly with the trajectories of both the camera and the moving agents within a novel bundle adjustment proposal. 
The 3D bounding boxes of the objects are also estimated and loosely optimized within a fixed temporal window. 
We demonstrate that tracking dynamic objects does not only provide rich clues for scene understanding but is also beneficial for  camera tracking. 

The project code  will be released upon acceptance.
\end{abstract}


%%%%%%%%%%%%%%%%%%%%%%%%%%%%%%%%%%%%%%%%%%%%%%%%%%%%%%%%%%%%%%%%%%%%%%%%%%%%%%%%
\section{Introduction}
\label{sec: introduction}

Visual Simultaneous Localization and Mapping (SLAM) is the problem of creating a map of an unknown environment and estimating the robot pose within such map, only from the data streams of its on-board cameras. 
Most SLAM approaches assume a static scene and can only handle small fractions of dynamic content by labeling them as outliers to such static model~\cite{mur2017orb, engel2017direct, forster2014svo}. 
Whilst the static premise holds for some robotic applications, it limits its use in populated situations for autonomous driving, service robots or AR/VR. 

The problem of dealing with dynamic objects in SLAM has been widely targeted in recent years. 
The biggest part of the literature tackles this problem by detecting moving regions within the observed scene and rejecting such areas for the SLAM problem~\cite{bescos2018dynaslam, sun2017improving, li2017rgb, xiao2019dynamic}. 
Some works process the image streams outside of the localization pipeline by translating the images that show dynamic content into realistic images with only static content~\cite{bescos2019empty, bevsic2020dynamic}. 
On the other hand, a small but growing part of the robotics community has addressed this issue by incorporating the dynamics of moving objects into the problem~\cite{li2018stereo, yang2019cubeslam, huang2020clustervo, zhang2020vdo}. 
Whereas the two first groups mostly focus on achieving an accurate ego-motion estimation from the static scene, the goal of the latter group is twofold: they do not only solve the SLAM problem but also provide information about the poses of other dynamic agents. 

\begin{figure} [t]
    \centering
    \subfloat[\label{fig: teaser_rgb}The 3D bounding box and the speed of the objects are inferred in the image. 
    Static and dynamic key points are in green and red respectively.]{\includegraphics[width=0.98\linewidth]{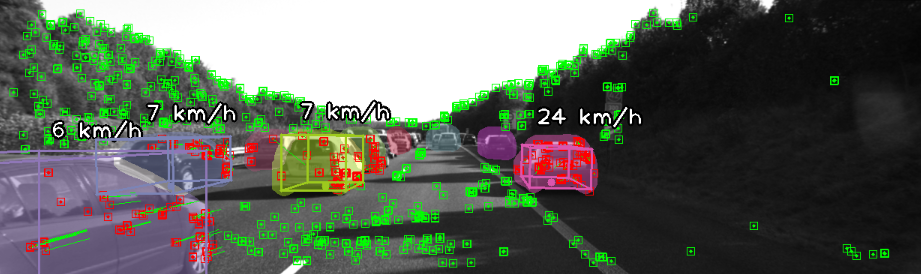}}
    \\
    \subfloat[\label{fig: teaser_map}Joint estimation of the camera ego motion (green car), the sparse static 3D map (black points) and the trajectories of the dynamic objects. 
    The cyan key frames allow to optimize the map dynamic structure, whereas the blue ones only optimize the camera pose and the static structure.]{\includegraphics[width=0.98\linewidth]{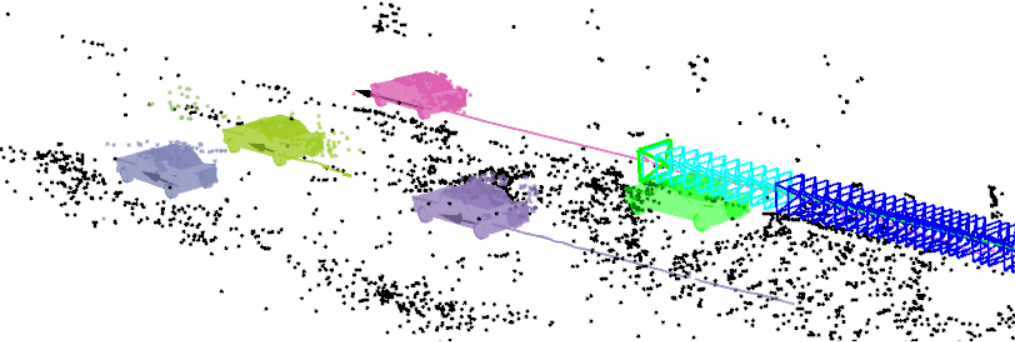}}
    \caption{Qualitative results with the KITTI tracking dataset.}
    \label{fig: teaser_image}
\end{figure}

Understanding surrounding dynamic objects is of crucial importance for the frontier requirements of emerging applications within AR/VR or autonomous systems navigation. 
Whereas it is tolerable to rule out minor movements in a quasi-static environment, most scenarios including autonomous driving, multi-robot collaboration and AR/VR require explicit motion information of the surroundings to aid in decision-making and scene understanding. 
For example, in virtual reality, dynamic objects need to be explicitly tracked to allow the interaction of virtual objects with moving instances in the real world.
In autonomous driving scenarios, a car must not only localize itself but must also reliably perceive other vehicles and passers-by to avoid collisions.

The vast majority of the literature that specifically addresses this issue detect moving objects and track them separately from the SLAM formulation by using traditional multi-target tracking approaches \cite{wang2003online, wangsiripitak2009avoiding, rogers2010slam, barsan2018robust, rosinol20203d}. 
Their accuracy highly depends on the camera pose estimation, which is more susceptible to failure in complex dynamic environments where the presence of reliable static structure is not guaranteed. 
In recent  years, the robotics community has made its first steps towards addressing dynamic objects tracking jointly with visual SLAM, adding an extra layer of complexity to the problem. 
These systems are often tailored for special use cases and multiple priors are exploited to constraint the space solutions: road planar structure and planar object movement in driving scenarios, or even the use of object 3D models.

At this point, we take the opportunity to introduce DynaSLAM~II. 
DynaSLAM~II is an open-source stereo and \mbox{RGB-D} SLAM system for dynamic environments which simultaneously estimates the poses of the camera, the map and the trajectories of the scene moving objects. 
We propose a bundle adjustment solution that tightly optimizes the scene structure, the camera poses and the objects trajectories in a local-temporal window. 
The bounding boxes of the objects are also optimized in a decoupled formulation that allows to estimate the dimensions and the 6 DoF poses of the objects without being geared to any particular use case. 
We provide and exhaustive evaluation and comparison on the KITTI dataset~\cite{geiger2013vision}, validating our proposal. 
An example of the output of our system DynaSLAM~II can be seen in Fig.~\ref{fig: teaser_image}. 

\section{Related Work}
\label{sec: related_work}

The traditional manner of addressing 3D multi-object tracking implies detecting and tracking the moving objects separately from the SLAM formulation~\cite{wang2003online, wangsiripitak2009avoiding, rogers2010slam, barsan2018robust, rosinol20203d}.
Among them, Wang \textit{et~al.}~\cite{wang2003online} derived the Bayes formula of the SLAM with tracking of moving objects and provided a solid basis for understanding and solving this problem. 
Wangsiripitak \textit{et~al.}~\cite{wangsiripitak2009avoiding} proposed the parallel implementation of SLAM with a 3D object tracker: the SLAM provides the tracker with information to register map objects, and the tracker allows to mark features on objects. 
Rogers \textit{et~al.}~\cite{rogers2010slam} applied an EM technique to a graph based SLAM approach and allowed landmarks to be dynamic. 
More recently, Barsan \textit{et~al.}~\cite{barsan2018robust} presented a stereo-based dense mapping algorithm for urban environments that simultaneously reconstructs the static background and the moving objects. 
There is a new work by Rosinol \textit{et~al.}~\cite{rosinol20203d} that reconciles visual-inertial SLAM and dense mesh tracking, focusing mostly on humans, that shows impressive results in simulation. 
The main drawback of these approaches is that their accuracy is highly correlated with the estimation of the camera position. 
That is, if the camera pose estimation fails, which is quite likely in complex dynamic environments, multi-object tracking also fails directly.

The idea of simultaneously estimating camera motion and multiple moving objects motion originated from the SLAMMOT work~\cite{wang2007simultaneous}. 
They established a mathematical framework to integrate a filtering-based SLAM and moving object tracking and demonstrated that it satisfied both navigation and safety requirements in autonomous driving. 
Later on, works that use \mbox{RGB-D} cameras followed up this idea to densely reconstruct static indoors scenes along with moving objects using pixel-wise instance segmentation, showing impressive results~\cite{runz2017co, runz2018maskfusion, xu2019mid}. 
Since it is of crucial importance for dense approaches to obtain a highly accurate segmentation mask, the works Mask-Fusion~\cite{runz2018maskfusion} and MID-Fusion~\cite{xu2019mid} refine it by assuming that human-made objects are largely convex. 

Among the feature-based approaches, as is ours, few aim to merge information from static and dynamic objects into a single framework to boost  estimation accuracy. 
Henein \textit{et~al.}~\cite{henein2018exploiting} were among the first ones to tightly combine the problems of tracking dynamic objects and the camera ego motion. 
However, they only reported experiments on synthetic data showing limited real results. 
Li \textit{et~al.}~\cite{li2018stereo} use a CNN trained in an end-to-end manner to estimate the 3D pose and dimensions of cars, which is further refined together with camera poses. 
The use of data-driven approaches often provides excellent accuracy in 6 DoF object pose estimation, but also a loss of generality and thus, they can only track cars. 
Huge amounts of data would be required to track generic objects with their approach. 
The authors of CubeSLAM~\cite{yang2019cubeslam} showed impressive results with only a monocular camera by making use of a 3D bounding box proposal generation based on 2D bounding boxes and vanishing points. 
They assume that objects have a constant velocity  within a hard-coded duration time interval and exploit object priors such as car sizes,  road  structure and  planar non-holonomic object wheel motion models. 
Moreover, they only track objects whose 3D bounding box is observable, \textit{i.e.}, only once two or more faces of the \textit{cuboid-shape} object are seen. 
ClusterSLAM~\cite{huang2019clusterslam} proposes a SLAM back end with no scene priors to discover individual rigid bodies and compute their motions in dynamic environments. 
Since it acts as a back end instead of as a full system, its performance relies heavily on the landmark tracking and association quality. 
The same authors recently developed the full system ClusterVO~\cite{huang2020clustervo}, which models the object points with a probability of object belonging to deal with segmentation inaccuracies. 
Given that they assume no priors, they obtain good tracking results in indoor and outdoor scenes, but with an inaccurate estimation of the 3D bounding boxes. 
VDO-SLAM~\cite{zhang2020vdo} is a recent work that uses dense optical flow to maximise the number of tracked points on moving objects. 
They implement a bundle adjustment with cameras, objects and points that gives good results but is computationally complex to run in real time.

In light of these advances, it is apparent that the feature-based SLAM community is searching for the best optimization formulation to combine cameras, objects and structure points. 
In our proposal, we use a tightly-coupled bundle adjustment formulation with new measurements between cameras, objects and points giving special attention to its computationally complexity and number of parameters involved without introducing hard-coded priors. 
For this, we integrate instance semantic priors together with sparse image features. 
This formulation allows the estimation of both the camera, the map structure and the dynamic objects to be mutually beneficial at a low computational cost. 
On the other hand, part of the current literature focuses on the estimation of the point cloud structure of dynamic objects and of the trajectory of a random object reference~\cite{huang2019clusterslam, huang2020clustervo, zhang2020vdo}, whereas another part of the literature seeks to find a common reference for objects of the same class as well as a more informative occupancy volume~\cite{li2018stereo, yang2019cubeslam}. 
We intend to carry out these two tasks independently in order to leverage the benefits of both and not suffer their disadvantages.

\section{Method}
\label{sec: method}

DynaSLAM~II builds on the popular ORB-SLAM2~\cite{mur2017orb}.  It takes synchronized and calibrated stereo/\mbox{RGB-D} images as input, and outputs the camera and the dynamic objects poses for each frame, as well as a spatial/temporal map containing the dynamic objects. 
For each incoming frame, pixel-wise semantic segmentation is computed and ORB features~\cite{rublee2011orb} are extracted and matched across stereo image pairs. 
We first associate the static and dynamic features with the ones from the previous frame and the map assuming a constant velocity motion for both the camera and the observed objects. 
Object instances are then matched based on the dynamic feature correspondences. 
The static matches are used to estimate the initial camera pose, and the dynamic ones yield the object's $\text{SE}(3)$ transform. 
Finally, the camera and objects trajectories, as well as the objects bounding boxes and 3D points are optimized over a sliding window with marginalization and a soft smooth motion prior. 
The different contributions and building blocks of DynaSLAM~II are explained in the following subsections.

\subsection{Notation}

We will use the following notation: a stereo/\mbox{RGB-D} camera $i$ has a pose $\mathbf{T}_\mathtt{CW}^i \in \text{SE}(3)$ in the world coordinates $\mathbf{W}$ at time $i$ (see Fig.~\ref{fig: notation}). 
The camera $i$ observes 
\begin{enumerate*}
\item static 3D map points $\mathbf{x}^l_\mathtt{W} \in \mathbb{R}^3 $ and
\item dynamic objects with pose $\mathbf{T}_\mathtt{WO}^{k,i} \in \text{SE}(3)$ and linear and angular velocity $\mathbf{v}^k_i, \mathbf{w}^k_i \in \mathbb{R}^3$ at time $i$, all in object coordinates. 
\end{enumerate*}
Each observed object $k$ contains dynamic objects points $\mathbf{x}^{j,k}_\mathtt{O} \in \mathbb{R}^3$. 

%
% The set of object points $\mathcal{OP}_k$ belonging to the same instance lets us define the dynamic object $k \in \mathcal{O}_i$ at time $i$, which has a linear and angular velocity in object coordinates $\mathbf{v}^k_i, \mathbf{w}^k_i \in \mathbb{R}^3$ respectively, and a world coordinates pose $\mathbf{T}_\mathtt{OW}^{k,i} \in \text{SE}(3)$.

\subsection{Objects Data Association}

For each incoming frame, pixel-wise semantic segmentation is computed and ORB features~\cite{rublee2011orb} are extracted and matched across stereo image pairs. 
If an instance belongs to a dynamic class (vehicles, pedestrians and animals) and contains a high number of new nearby key points, an object is created. 
The key points are then assigned to the instance and the corresponding object. 
We first associate the static features with the ones from the previous frame and the map to initially estimate the camera pose. 
Next, dynamic features are associated with the dynamic points from the local map in two different ways: 
\begin{enumerate*}[label=(\alph*)]
    \item if the velocity of the map objects is known, the matches are searched by reprojection assuming an inter-frame constant velocity motion, 
    \item if the objects velocity is not initialized or not enough matches are found following (a), we constraint the brute force matching to those features that belong to the most overlapping instance within consecutive frames.
\end{enumerate*} 
Note that our framework handles occlusions since current key points are matched to map objects instead of to the previous frame objects.
A higher level association between instances and objects is also required. 
If most of the key points assigned to a new object are matched with points belonging to one map object, the two objects are attributed the same track id. 
Also, to render this high level association more robust, a parallel instance-to-instance matching is performed based on the IoU (Intersection over Union) of the CNN instances 2D bounding boxes.

The SE(3) pose of the first object of a track is initialized with the center of mass of the 3D points and with the identity rotation. 
To predict the poses of further objects from a track, we use a constant velocity motion model and refine the object pose estimate by minimizing the reprojection error. 

The common formulation of the reprojection error in multi-view geometry problems for a camera $i$ with pose $\mathbf{T}_\mathtt{CW}^{i} \in \text{SE}(3)$ and a 3D map point $l$ with homogeneous coordinates $\mathbf{\bar{x}}_\mathtt{W}^{l} \in \mathbb{R}^4$ in the reference $\mathcal{W}$ with a stereo key point correspondence $\mathbf{u}_{i}^{l} = [u, v, u_R] \in \mathbb{R}^3$ is

\begin{equation}\label{eqn: repr_error}
\mathbf{e}_{\mathbf{repr}}^{i,l} = \mathbf{u}_{i}^{l} -  \pi_{i} ( \mathbf{T}_{\mathtt{CW}}^{i} \mathbf{\bar{x}}_{\mathtt{W}}^{l} ) ,
\end{equation}

where $\pi_{i}$ is the reprojection function for a rectified stereo/\mbox{RGB-D} camera that projects a 3D homogeneous point in the camera coordinates into the camera frame pixel. 
Unlike this formulation, which is valid for static representations, we propose to restate the reprojection error as

\begin{equation}\label{eqn: repr_error_wo}
\mathbf{e}_{\mathbf{repr}}^{i,j,k} = \mathbf{u}_{i}^{j} -  \pi_{i} ( \mathbf{T}_{\mathtt{CW}}^{i}\mathbf{T}_{\mathtt{WO}}^{k,i} \mathbf{\bar{x}}_{\mathtt{O}}^{j,k} ) ,
\end{equation}

where $\mathbf{T}_\mathtt{WO}^{k,i} \in \text{SE}(3)$ is the inverse pose of the object $k$ in the world coordinates when the camera $i$ is observing it, and $\mathbf{\bar{x}}_\mathtt{O}^{j,k} \in \mathbb{R}^4$ represents the 3D homogeneous coordinates of the point $j$ in its object reference $k$ with observation in the camera $i$ $\mathbf{u}_{i}^{j} \in \mathbb{R}^3$. 
This formulation enables us to optimize either jointly the poses of the cameras and of the different moving objects, as well as the positions of their 3D points.

\newcommand\solidrule[1][1cm]{\rule[0.5ex]{#1}{.4pt}}

\newcommand\dashedrule[1][1mm]{\mbox{%
  \solidrule[#1]\hspace{#1}\solidrule[#1]\hspace{#1}\solidrule[#1]}}

\newcommand\dashdottedrule[2][1=0.5mm, 2=2mm]{\mbox{%
  \solidrule[#1]\hspace{0.5mm}\solidrule[#2]\hspace{0.5mm}\solidrule[#1]\hspace{0.5mm}\solidrule[#2]\hspace{0.5mm}\solidrule[#1]}}

% [1=defaultFirstArg, 3=defaultThirdArg]{#1~#2~#3~#4}

\begin{figure}[t]
    \centering
    \includegraphics[width=0.90
    \linewidth]{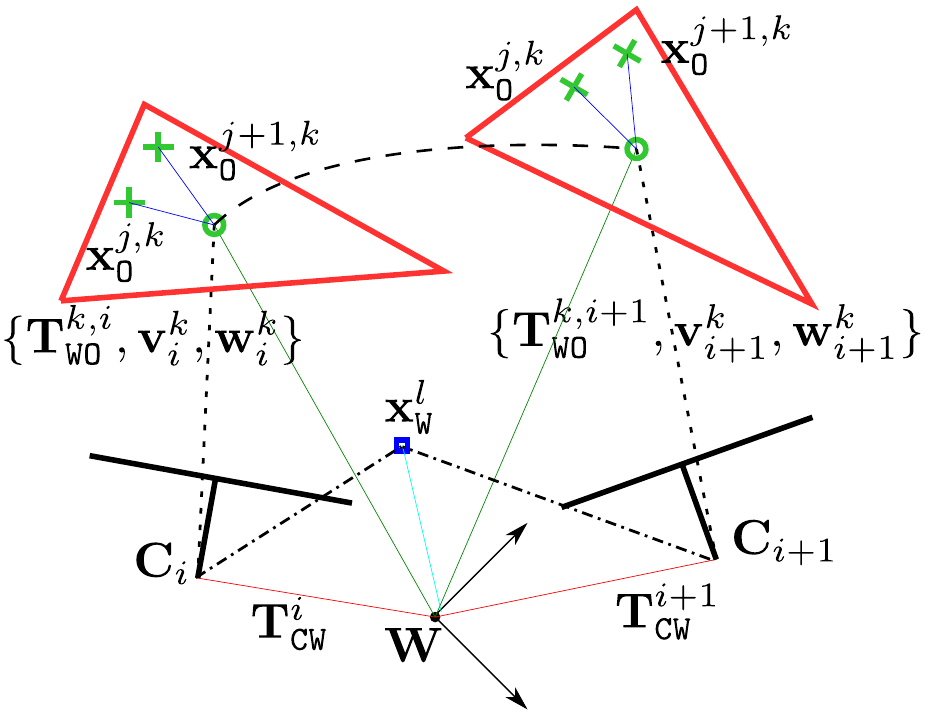}
    \caption{Notation used to model the dynamic structure. 
    The cameras $i$ and $i+1$ observe the dynamic object $k$ (\protect\dashedrule[0.5mm]) and the static structure (\protect\dashdottedrule[0.25mm]{1.25mm}). 
    The objects with poses $\mathbf{T}_\mathtt{WO}^{k,i}$ and $\mathbf{T}_\mathtt{WO}^{k,i+1}$ are the same moving body at consecutive observations (\protect\dashedrule[1.25mm]).}
    \label{fig: notation}
\end{figure}

\subsection{Object-Centric Representation}

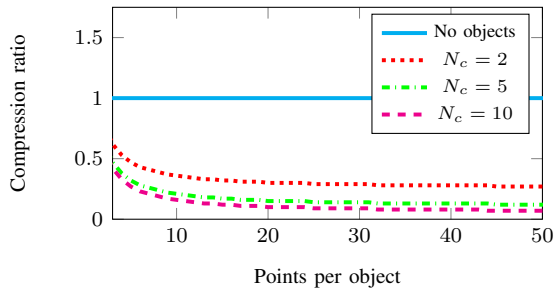
\begin{figure}[t]
\centering 
\begin{tikzpicture}

\begin{axis}[%
width=2.25in,
height=1.11in,
scale only axis,
xmin=3,
xmax=50,
xlabel style={font=\footnotesize},
xlabel={Points per object},
ymin=0,
ymax=1.75,
ylabel style={font=\footnotesize},
ylabel={Compression ratio},
axis background/.style={fill=white},
tick label style={font=\footnotesize},
legend style={legend pos=north east, font=\scriptsize}
%legend style={legend pos=north east,draw=black,fill=white,legend cell align=left, font=\small}
%xmajorgrids,
%ymajorgrids
]
\addplot [color=cyan, line width=1.5pt]
  table[row sep=crcr]{%
    1.00 1.00 \\ 
     2.00 1.00 \\ 
     3.00 1.00 \\ 
     4.00 1.00 \\ 
     5.00 1.00 \\ 
     6.00 1.00 \\ 
     7.00 1.00 \\ 
     8.00 1.00 \\ 
     9.00 1.00 \\ 
     10.00 1.00 \\ 
     11.00 1.00 \\ 
     12.00 1.00 \\ 
     13.00 1.00 \\ 
     14.00 1.00 \\ 
     15.00 1.00 \\ 
     16.00 1.00 \\ 
     17.00 1.00 \\ 
     18.00 1.00 \\ 
     19.00 1.00 \\ 
     20.00 1.00 \\ 
     21.00 1.00 \\ 
     22.00 1.00 \\ 
     23.00 1.00 \\ 
     24.00 1.00 \\ 
     25.00 1.00 \\ 
     26.00 1.00 \\ 
     27.00 1.00 \\ 
     28.00 1.00 \\ 
     29.00 1.00 \\ 
     30.00 1.00 \\ 
     31.00 1.00 \\ 
     32.00 1.00 \\ 
     33.00 1.00 \\ 
     34.00 1.00 \\ 
     35.00 1.00 \\ 
     36.00 1.00 \\ 
     37.00 1.00 \\ 
     38.00 1.00 \\ 
     39.00 1.00 \\ 
     40.00 1.00 \\ 
     41.00 1.00 \\ 
     42.00 1.00 \\ 
     43.00 1.00 \\ 
     44.00 1.00 \\ 
     45.00 1.00 \\ 
     46.00 1.00 \\ 
     47.00 1.00 \\ 
     48.00 1.00 \\ 
     49.00 1.00 \\ 
     50.00 1.00 \\ 
     51.00 1.00 \\ 
     52.00 1.00 \\ 
     53.00 1.00 \\ 
     54.00 1.00 \\ 
     55.00 1.00 \\ 
     56.00 1.00 \\ 
     57.00 1.00 \\ 
     58.00 1.00 \\ 
     59.00 1.00 \\ 
     60.00 1.00 \\ 
     61.00 1.00 \\ 
     62.00 1.00 \\ 
     63.00 1.00 \\ 
     64.00 1.00 \\ 
     65.00 1.00 \\ 
     66.00 1.00 \\ 
     67.00 1.00 \\ 
     68.00 1.00 \\ 
     69.00 1.00 \\ 
     70.00 1.00 \\ 
     71.00 1.00 \\ 
     72.00 1.00 \\ 
     73.00 1.00 \\ 
     74.00 1.00 \\ 
     75.00 1.00 \\ 
     76.00 1.00 \\ 
     77.00 1.00 \\ 
     78.00 1.00 \\ 
     79.00 1.00 \\ 
     80.00 1.00 \\ 
     81.00 1.00 \\ 
     82.00 1.00 \\ 
     83.00 1.00 \\ 
     84.00 1.00 \\ 
     85.00 1.00 \\ 
     86.00 1.00 \\ 
     87.00 1.00 \\ 
     88.00 1.00 \\ 
     89.00 1.00 \\ 
     90.00 1.00 \\ 
     91.00 1.00 \\ 
     92.00 1.00 \\ 
     93.00 1.00 \\ 
     94.00 1.00 \\ 
     95.00 1.00 \\ 
     96.00 1.00 \\ 
     97.00 1.00 \\ 
     98.00 1.00 \\ 
     99.00 1.00 \\ 
     100.00 1.00 \\ 
};

\addlegendentry{No objects}

\addplot [color=red, dotted, line width=1.5pt]
  table[row sep=crcr]{%
    1.00 1.35 \\
     2.00 0.80 \\
     3.00 0.62 \\
     4.00 0.53 \\
     5.00 0.47 \\
     6.00 0.43 \\
     7.00 0.41 \\
     8.00 0.39 \\
     9.00 0.37 \\
     10.00 0.36 \\
     11.00 0.35 \\
     12.00 0.34 \\
     13.00 0.33 \\
     14.00 0.33 \\
     15.00 0.32 \\
     16.00 0.32 \\
     17.00 0.31 \\
     18.00 0.31 \\
     19.00 0.31 \\
     20.00 0.30 \\
     21.00 0.30 \\
     22.00 0.30 \\
     23.00 0.30 \\
     24.00 0.30 \\
     25.00 0.29 \\
     26.00 0.29 \\
     27.00 0.29 \\
     28.00 0.29 \\
     29.00 0.29 \\
     30.00 0.29 \\
     31.00 0.29 \\
     32.00 0.28 \\
     33.00 0.28 \\
     34.00 0.28 \\
     35.00 0.28 \\
     36.00 0.28 \\
     37.00 0.28 \\
     38.00 0.28 \\
     39.00 0.28 \\
     40.00 0.28 \\
     41.00 0.28 \\
     42.00 0.28 \\
     43.00 0.28 \\
     44.00 0.28 \\
     45.00 0.27 \\
     46.00 0.27 \\
     47.00 0.27 \\
     48.00 0.27 \\
     49.00 0.27 \\
     50.00 0.27 \\
     51.00 0.27 \\
     52.00 0.27 \\
     53.00 0.27 \\
     54.00 0.27 \\
     55.00 0.27 \\
     56.00 0.27 \\
     57.00 0.27 \\
     58.00 0.27 \\
     59.00 0.27 \\
     60.00 0.27 \\
     61.00 0.27 \\
     62.00 0.27 \\
     63.00 0.27 \\
     64.00 0.27 \\
     65.00 0.27 \\
     66.00 0.27 \\
     67.00 0.27 \\
     68.00 0.27 \\
     69.00 0.27 \\
     70.00 0.27 \\
     71.00 0.27 \\
     72.00 0.27 \\
     73.00 0.27 \\
     74.00 0.26 \\
     75.00 0.26 \\
     76.00 0.26 \\
     77.00 0.26 \\
     78.00 0.26 \\
     79.00 0.26 \\
     80.00 0.26 \\
     81.00 0.26 \\
     82.00 0.26 \\
     83.00 0.26 \\
     84.00 0.26 \\
     85.00 0.26 \\
     86.00 0.26 \\
     87.00 0.26 \\
     88.00 0.26 \\
     89.00 0.26 \\
     90.00 0.26 \\
     91.00 0.26 \\
     92.00 0.26 \\
     93.00 0.26 \\
     94.00 0.26 \\
     95.00 0.26 \\
     96.00 0.26 \\
     97.00 0.26 \\
     98.00 0.26 \\
     99.00 0.26 \\
     100.00 0.26 \\
};

\addlegendentry{$N_c = 2$}
    
\addplot [color=green, dash dot, line width=1.5pt]
  table[row sep=crcr]{% 
    1.00 1.20 \\
     2.00 0.65 \\
     3.00 0.47 \\
     4.00 0.38 \\
     5.00 0.32 \\
     6.00 0.28 \\
     7.00 0.26 \\
     8.00 0.24 \\
     9.00 0.22 \\
     10.00 0.21 \\
     11.00 0.20 \\
     12.00 0.19 \\
     13.00 0.18 \\
     14.00 0.18 \\
     15.00 0.17 \\
     16.00 0.17 \\
     17.00 0.16 \\
     18.00 0.16 \\
     19.00 0.16 \\
     20.00 0.15 \\
     21.00 0.15 \\
     22.00 0.15 \\
     23.00 0.15 \\
     24.00 0.15 \\
     25.00 0.14 \\
     26.00 0.14 \\
     27.00 0.14 \\
     28.00 0.14 \\
     29.00 0.14 \\
     30.00 0.14 \\
     31.00 0.14 \\
     32.00 0.13 \\
     33.00 0.13 \\
     34.00 0.13 \\
     35.00 0.13 \\
     36.00 0.13 \\
     37.00 0.13 \\
     38.00 0.13 \\
     39.00 0.13 \\
     40.00 0.13 \\
     41.00 0.13 \\
     42.00 0.13 \\
     43.00 0.13 \\
     44.00 0.13 \\
     45.00 0.12 \\
     46.00 0.12 \\
     47.00 0.12 \\
     48.00 0.12 \\
     49.00 0.12 \\
     50.00 0.12 \\
     51.00 0.12 \\
     52.00 0.12 \\
     53.00 0.12 \\
     54.00 0.12 \\
     55.00 0.12 \\
     56.00 0.12 \\
     57.00 0.12 \\
     58.00 0.12 \\
     59.00 0.12 \\
     60.00 0.12 \\
     61.00 0.12 \\
     62.00 0.12 \\
     63.00 0.12 \\
     64.00 0.12 \\
     65.00 0.12 \\
     66.00 0.12 \\
     67.00 0.12 \\
     68.00 0.12 \\
     69.00 0.12 \\
     70.00 0.12 \\
     71.00 0.12 \\
     72.00 0.12 \\
     73.00 0.12 \\
     74.00 0.11 \\
     75.00 0.11 \\
     76.00 0.11 \\
     77.00 0.11 \\
     78.00 0.11 \\
     79.00 0.11 \\
     80.00 0.11 \\
     81.00 0.11 \\
     82.00 0.11 \\
     83.00 0.11 \\
     84.00 0.11 \\
     85.00 0.11 \\
     86.00 0.11 \\
     87.00 0.11 \\
     88.00 0.11 \\
     89.00 0.11 \\
     90.00 0.11 \\
     91.00 0.11 \\
     92.00 0.11 \\
     93.00 0.11 \\
     94.00 0.11 \\
     95.00 0.11 \\
     96.00 0.11 \\
     97.00 0.11 \\
     98.00 0.11 \\
     99.00 0.11 \\
     100.00 0.11 \\
};

\addlegendentry{$N_c = 5$}

\addplot [color=magenta, dashed, line width=1.5pt]
  table[row sep=crcr]{% 
    1.00 1.15 \\
     2.00 0.60 \\
     3.00 0.42 \\
     4.00 0.33 \\
     5.00 0.27 \\
     6.00 0.23 \\
     7.00 0.21 \\
     8.00 0.19 \\
     9.00 0.17 \\
     10.00 0.16 \\
     11.00 0.15 \\
     12.00 0.14 \\
     13.00 0.13 \\
     14.00 0.13 \\
     15.00 0.12 \\
     16.00 0.12 \\
     17.00 0.11 \\
     18.00 0.11 \\
     19.00 0.11 \\
     20.00 0.10 \\
     21.00 0.10 \\
     22.00 0.10 \\
     23.00 0.10 \\
     24.00 0.10 \\
     25.00 0.09 \\
     26.00 0.09 \\
     27.00 0.09 \\
     28.00 0.09 \\
     29.00 0.09 \\
     30.00 0.09 \\
     31.00 0.09 \\
     32.00 0.08 \\
     33.00 0.08 \\
     34.00 0.08 \\
     35.00 0.08 \\
     36.00 0.08 \\
     37.00 0.08 \\
     38.00 0.08 \\
     39.00 0.08 \\
     40.00 0.08 \\
     41.00 0.08 \\
     42.00 0.08 \\
     43.00 0.08 \\
     44.00 0.07 \\
     45.00 0.07 \\
     46.00 0.07 \\
     47.00 0.07 \\
     48.00 0.07 \\
     49.00 0.07 \\
     50.00 0.07 \\
     51.00 0.07 \\
     52.00 0.07 \\
     53.00 0.07 \\
     54.00 0.07 \\
     55.00 0.07 \\
     56.00 0.07 \\
     57.00 0.07 \\
     58.00 0.07 \\
     59.00 0.07 \\
     60.00 0.07 \\
     61.00 0.07 \\
     62.00 0.07 \\
     63.00 0.07 \\
     64.00 0.07 \\
     65.00 0.07 \\
     66.00 0.07 \\
     67.00 0.07 \\
     68.00 0.07 \\
     69.00 0.07 \\
     70.00 0.07 \\
     71.00 0.07 \\
     72.00 0.07 \\
     73.00 0.07 \\
     74.00 0.06 \\
     75.00 0.06 \\
     76.00 0.06 \\
     77.00 0.06 \\
     78.00 0.06 \\
     79.00 0.06 \\
     80.00 0.06 \\
     81.00 0.06 \\
     82.00 0.06 \\
     83.00 0.06 \\
     84.00 0.06 \\
     85.00 0.06 \\
     86.00 0.06 \\
     87.00 0.06 \\
     88.00 0.06 \\
     89.00 0.06 \\
     90.00 0.06 \\
     91.00 0.06 \\
     92.00 0.06 \\
     93.00 0.06 \\
     94.00 0.06 \\
     95.00 0.06 \\
     96.00 0.06 \\
     97.00 0.06 \\
     98.00 0.06 \\
     99.00 0.06 \\
     100.00 0.06 \\
};
     
\addlegendentry{$N_c = 10$}

\end{axis}
\end{tikzpicture}%
\caption{\label{fig: motivation_objects} Relationship between the number of parameters that is required when objects are used and when points belonging to objects are tracked independently (No~objects).}
\end{figure}

Given the extra complexity and mainly the extra number of parameters that the task of tracking moving objects implies on top of the SLAM ones, it is of high importance to keep this number as reduced as possible to maintain a real-time performance. 
Modeling dynamic points as repeated 3D points by forming independent point clouds as in usual dynamic SLAM implementations results in a prohibitive amount of parameters. 
Given a set of $N_c$ cameras, $N_o$ dynamic objects with $N_{op}$ 3D points each observed in all cameras, the number of parameters needed to track dynamic objects becomes $N = 6 N_c + N_c \times N_o \times 3 N_{op}$ as opposed to $N = 6 N_c + N_o \times 3 N_{op}$ in conventional static SLAM representations. 
This number of parameters becomes prohibitive for long --and not so long-- operations and deployment. 
If the concept of objects is introduced, 3D object points become unique and can be referred to their dynamic object. 
Therefore it is the pose of the object that is modelled along time and the number of required parameters shifts to $N' = 6 N_c + N_c \times 6 N_o + N_o \times 3 N_{op}$. 
Fig.~\ref{fig: motivation_objects} shows the parameter compression ratio defined as $\frac{N'}{N}$ for 10 objects. 
This modelling of dynamic objects and points brings great savings in the number of utilized parameters.

\subsection{Bundle Adjustment with Objects}

\begin{figure}[t]
    \centering
    \includegraphics[width=0.87
    \linewidth]{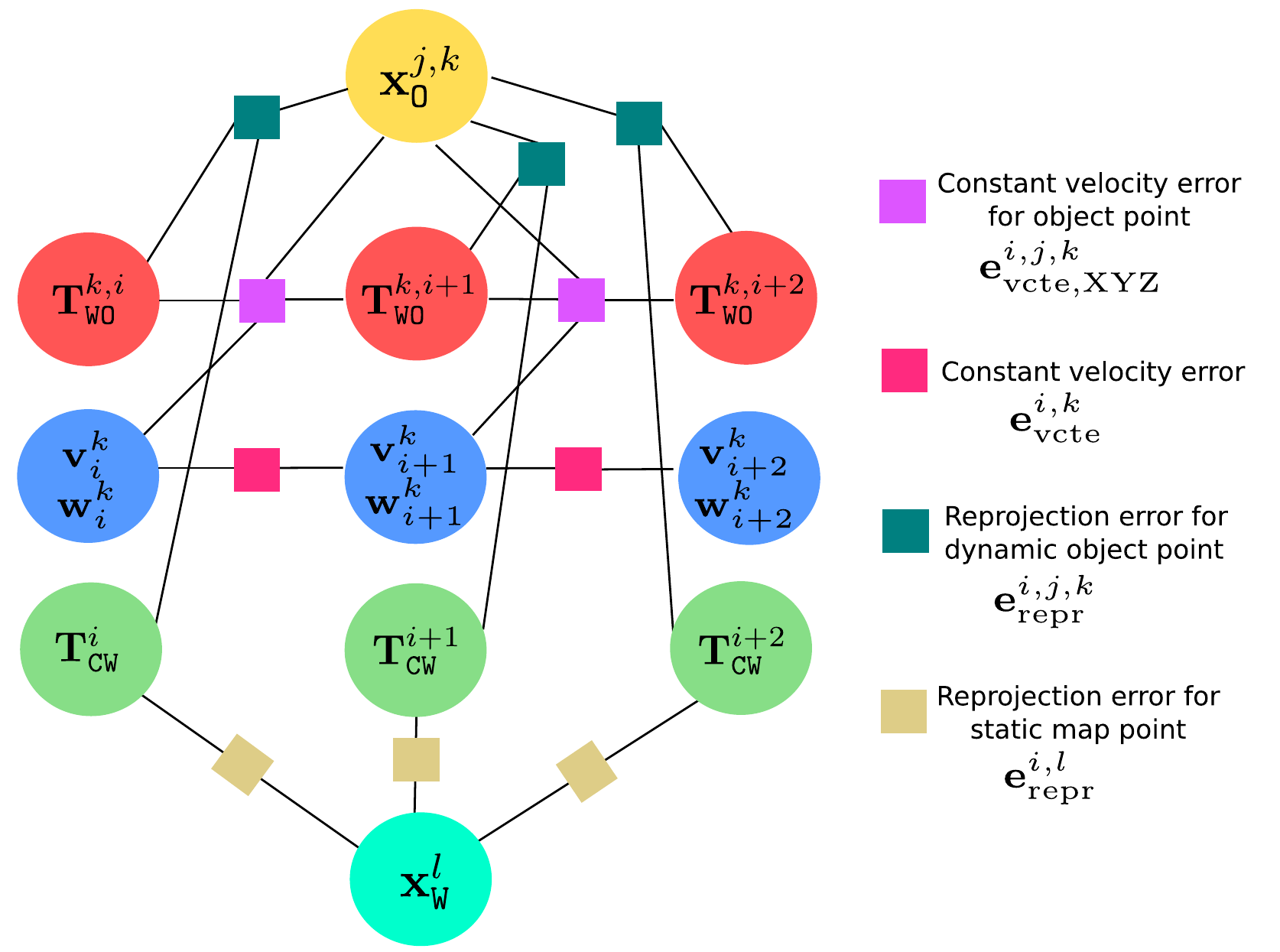}
    \caption{BA factor graph representation with dynamic objects.}
    \label{fig: factor_graph}
\end{figure}

Bundle Adjustment (BA) is known to provide accurate estimates of camera poses and sparse geometrical reconstruction, given a strong network of matches and good initial guesses. 
We hypothesize that BA might bring similar benefits if object poses are also jointly optimized (Fig.~\ref{fig: factor_graph}). 
The static map point 3D locations $\mathbf{\bar{x}}_\mathtt{W}^{l}$ and camera poses $\mathbf{T}_\mathtt{CW}^{i}$ are optimized by minimizing the reprojection error with respect to the matched key points $\mathbf{u}_{i}^{l}$ (Eqn.~\ref{eqn: repr_error}). 
Similarly, for dynamic representations, the object points $\mathbf{\bar{x}}_\mathtt{O}^{j,k}$, the camera poses $\mathbf{T}_\mathtt{CW}^{i}$ and the object poses $\mathbf{T}_\mathtt{WO}^{k,i}$ can be refined by minimizing the reprojection error formulation presented in Eqn.~\ref{eqn: repr_error_wo}.

In our implementation, a key frame can be inserted in the map for two different reasons: 
\begin{enumerate*}[label=(\alph*)]
    \item the camera tracking is weak, 
    \item the tracking of any scene object is weak. 
\end{enumerate*} 
The reasons for the former are the same ones than in ORB-SLAM2. 
The latter though happens if an object with a relatively large amount of features has few points tracked in the current frame. 
In this case, a key frame is inserted and creates a new object and new object points. 
If the camera tracking is not weak, this key frame would not introduce new static map points, and if the rest of dynamic objects have a stable tracking, new objects for these tracks would not be created. 
As for the optimization, if a key frame is inserted only because the camera tracking is weak, the local BA optimizes the currently processed key frame, all the key frames connected to it in the covisibility graph, and all the map points seen by those key frames, following the implementation of ORB-SLAM2. 
Regarding dynamic data, if a key frame is inserted only because the tracking of an object is weak, the local BA optimizes the pose and velocity of this object and the camera along a temporal tail of 2 seconds together with its object points. 
Finally, if a key frame is inserted because the tracking of both the camera and objects are weak, camera poses, map structure, object poses, velocities and points are jointly optimized.

To avoid non-physically feasible object dynamics, a smooth trajectory is forced by assuming a constant velocity in consecutive observations. 
The linear and angular velocity of an object $k$ at observation $i$ are respectively denoted as $\textbf{v}^{k}_{i} \in \mathbb{R}^3$ and $\textbf{w}^{k}_{i} \in \mathbb{R}^3$.  We define the following error term:

\begin{equation}\label{eqn: vcte}
    \mathbf{e}^{i,k}_{\mathbf{vcte}} = \left( \begin{array}{cc}
    \textbf{v}^{k}_{i+1} - \textbf{v}^{k}_{i} \\ 
    \textbf{w}^{k}_{i+1} - \textbf{w}^{k}_{i} 
    \end{array} \right) 
\end{equation}

An additional error term is needed to couple the object velocities with the object poses and their corresponding 3D points.
This term can be seen in Eqn.~\ref{eqn: vcte_XYZ}, where $\Delta \mathbf{T}_{\mathtt{O}_k}^{i,i+1}$ is the pose transformation that the object $k$ undergoes in the time interval $\Delta t_{i,i+1}$ between consecutive observations $i$ and~$i+1$. 

\begin{equation} \label{eqn: vcte_XYZ}
    \mathbf{e}_{\mathbf{vcte,XYZ}}^{i,j,k} = \left( \mathbf{T}_{\mathtt{WO}}^{k,i+1} - \mathbf{T}_{\mathtt{WO}}^{k,i} \Delta \mathbf{T}_{\mathtt{O}_k}^{i,i+1} \right) \mathbf{\bar{x}}_{\mathtt{O}}^{j,k}
\end{equation}

The term $\Delta \mathbf{T}_{\mathtt{O}_k}^{i,i+1}$ is defined from the linear and angular velocity of the object $k$ at time $i$ ($\mathbf{v}^k_i$ and $\mathbf{w}^k_i$) as in Eqn.~\ref{eqn: DeltaT}, where $\text{Exp}: \mathbb{R}^3 \rightarrow \text{SO}(3)$ is the exponential map for $\text{SO}(3)$. 

\begin{equation} \label{eqn: DeltaT}
\Delta \mathbf{T}_{\mathtt{O}_k}^{i,i+1} = \left( \begin{array}{cc}
    \text{Exp}(\mathbf{w}^k_i \Delta t_{i,i+1}) & \mathbf{v}^k_i \Delta t_{i,i+1} \\
    \mathbf{0}_{1 \times 3} & 1
\end{array} \right) 
\end{equation}

Finally, the following is our BA problem for a set of cameras in the optimizable local window $\mathcal{C}$ with each camera $i$ observing a set of map points $\mathcal{MP}_i$ and an object set $\mathcal{O}_i$ containing each object $k$ the set of object points $\mathcal{OP}_k$:

\begin{multline} \label{eqn: BA_static_dynamic}
    \min \limits_{\theta} \sum_{i\in\mathcal{C}} ( \sum_{l\in \mathcal{MP}_i} \rho \left.( \| \mathbf{e}_{\mathbf{repr}}^{i,l} \|^{2}_{\Sigma_{i}^{l}} \right.) + \sum_{k\in\mathcal{O}_i} ( \rho \left.( \| \mathbf{e}_{\mathbf{vcte}}^{i,k}\|^{2}_{\Sigma_{\Delta t}} \right.) \\ + \sum_{j\in \mathcal{OP}_k} \left.( \rho(\|\mathbf{e}_{\mathbf{repr}}^{i,j,k}\|^{2}_{\Sigma_{i}^{j}}) + \rho \left.( \| \mathbf{e}_{\mathbf{vcte,XYZ}}^{i,j,k}\|^{2}_{\Sigma_{\Delta t}} \right.) \right.) ) ),
\end{multline}

where $\rho$ is the robust Huber cost function to downweigh outlier correspondences and $\Sigma$ is the covariance matrix. 
In the case of the reprojection error $\Sigma$ is associated to the scale of the key point in the camera $i$ observing the points $l$ and $j$ respectively. 
For the two other error terms $\Sigma$ is associated to the time interval between two consecutive observations of an object, \textit{i.e.}, the longer time the more uncertainty there is about the constant velocity assumption. 
The parameters to be optimized are
$\theta = \{\mathbf{T}^{i}_{\mathtt{CW}}, \mathbf{T}^{k,i}_{\mathtt{WO}}, \mathbf{X}^l_\mathtt{W}, \mathbf{X}^{j,k}_\mathtt{O}, \mathbf{v}^{k}_{i}, \mathbf{w}^{k}_{i}\}$.
  
\begin{figure} [t]
\centering 
{\input{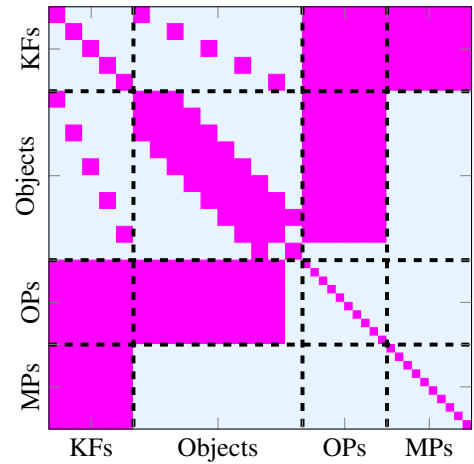}}
\caption{\label{fig: hessianBA} Hessian matrix for 5 key frames (KFs), 1 object with 10 object points (OPs) and 10 static map points (MPs).}
\end{figure}

Fig.~\ref{fig: hessianBA} shows the boolean Hessian matrix ($\mathbf{H}$) of the  problem described. 
The Hessian can be built from the Jacobian matrices associated to each edge in the factor graph. 
In order to have a non-zero $(i, j)$ block matrix, there must be an edge between $i$ and $j$ node in the factor graph.
Notice the difference in the sparsity patterns of the map points and the object points.
The size of the Hessian matrix is dominated by the number of map points $N_{mp}$ and object points, which in typical problems is several orders of magnitude larger than the number of cameras and objects. 
Applying Schur complement trick and solving the system has a run-time complexity of $\mathcal{O}(N_c^3 + N_c^2 N_{mp} + N_c N_o N_{op})$, where either the second or third term will dominate the cost depending on the number of static and dynamic points.

\subsection{Bounding Boxes}

Some approaches in the current literature would consider the tracking of dynamic objects to be complete with the contributions so far. 
That is, for every dynamic object in the scene we have an estimation of the trajectory of the centroid of its map points when it was first observed, as well as a point cloud representation. 
Examples of these works are ClusterSLAM \cite{huang2019clusterslam} and VDO-SLAM \cite{zhang2020vdo}. 
However, we believe that it is also of key importance to find a common spatial reference for objects of the same semantic class, as well as an estimate of their dimensions and space occupancy. 

Alternatively, the basis of CubeSLAM~\cite{yang2019cubeslam} and of the work by Li \textit{et~al.}~\cite{li2018stereo} is the discovery of object bounding boxes. 
Only once bounding boxes are discovered,  are objects tracked along frames. 
That is, if the camera viewing angle does not allow to estimate an object bounding box (partial view), the object tracking does not take place. 
Whereas this is not a problem for Li \textit{et~al.} because CNNs are by nature robust to partial views of objects, CubeSLAM struggles to initialize bounding boxes from views of occluded objects.

We propose to decouple the estimation of the trajectories and the bounding boxes of the dynamic objects. 
The former provides the system tracking with rich clues for ego-motion estimation, and the conjunction of both are useful to understand the dynamics of the surroundings. 
The output of the data association and the BA stages contains the camera poses, the structure of the static scene and the dynamic objects, and the 6 DoF trajectory of one point for each object. 
This one point is the center of mass of the object 3D points when it is first observed. 
Even though the center of mass changes along time with new points observations, the object pose that is tracked and optimized is referred to this first center of mass. 
To have a full understanding of the moving surroundings, it is of high importance to know the objects dimensions and space occupancy. 
Tackling the two problems independently allows to track dynamic objects from the first frame in which they appear independently of the camera-object view point.

We initialize an object bounding box by searching two perpendicular planes that fit roughly the majority of the object points. 
We hypothesize that even though objects are not always perfect cuboids, many can approximately fit a 3D bounding box. 
In the case in which only one plane is found, we add a prior on the rough dimensions of the non-observable direction that is related to the object class. 
This procedure is done within a RANSAC scheme: we choose the computed 3D bounding box that has the largest IoU of its image projection with the CNN 2D bounding box. 
This bounding box is computed once for every object track. 

To refine the bounding box dimensions and its pose relative to the object tracking reference, an image-based optimization is performed within a temporal window. 
This optimization seeks to minimize the distance between the 3D bounding box image projection and the CNN 2D bounding box prediction. 
Given that this problem is not observable for less than three views of an object, this is only performed once an object has at least three observing key frames. 
Also, to constraint the solution space in case the view of an object makes this problem non-observable (\textit{e.g.}, a car observed from the back), a soft prior about the object dimensions is included. 
Since this prior is tightly related to the object class, we believe that adding this soft prior does not mean a loss of generality.
Finally, the initial bounding box pose is set as a prior so that the optimization solution remains close.  

% One can see in Fig.~\ref{fig: bounding_boxes} the effect of the different errors and priors we use in our optimization. 
% %
% First, all three objects (Figs.~\ref{fig: bb_00},~\ref{fig: bb_01} and~\ref{fig: bb_02}) yield the same 2D image projection so, unless an object is observed in at least three frames at different view points, more constraints are required to render the problem observable. 
% %
% Second, forcing the 3D points to be near the found planes constraints most cases. 
% %
% Third, a prior about the object's dimensions is needed, otherwise, cases such as the ones in Figs.~\ref{fig: bb_01} and~\ref{fig: bb_02} are not fully constrained.

% \begin{figure} [h]
%     \centering
%     \null\hfill
%     \subfloat[\label{fig: bb_00}]{\includegraphics[width = 0.20\linewidth]{images/bounding_boxes_00.pdf}}
%     \hfill
%     \subfloat[\label{fig: bb_01}]{\includegraphics[width = 0.20\linewidth]{images/bounding_boxes_01.pdf}}
%     \hfill
%     \subfloat[\label{fig: bb_02}]{\includegraphics[width = 0.20\linewidth]{images/bounding_boxes_02.pdf}}
%     \hfill\null
%     \caption{Examples of camera and bounding box configurations.}
%     \label{fig: bounding_boxes}
% \end{figure}

\section{Experiments}
\label{sec: experiments}

In this section we detail the experiments carried out to test DynaSLAM~II. 
It is divided in two main blocks: one that assesses the effect of tracking objects on the estimation of camera motion  (Subsection~\ref{subsec: vo_experiments}), and one that analyzes the multi-object tracking performance (Subsection~\ref{subsec: mot_experiments}).

\begin{table*}[t]
\footnotesize
\begin{tabularx}{\linewidth}{@{}l*{13}{C}}
\toprule
\multicolumn{1}{c|}{seq} & \multicolumn{3}{c|}{ORB-SLAM2} & \multicolumn{3}{c||}{DynaSLAM} & \multicolumn{3}{c|}{VDO-SLAM} & \multicolumn{3}{c}{Ours}
\\
\multicolumn{1}{c|}{} & ATE~[m] & RPE\textsubscript{t}[m/f]~ & \multicolumn{1}{c|}{RPE\textsubscript{R}[\degree/f]} & ATE~[m] & RPE\textsubscript{t}[m/f] & \multicolumn{1}{c||}{RPE\textsubscript{R}[\degree/f]} & ATE~[m] & RPE\textsubscript{t}[m/f] & \multicolumn{1}{c|}{RPE\textsubscript{R}[\degree/f]} & ATE~[m] & RPE\textsubscript{t}[m/f] & \multicolumn{1}{c}{RPE\textsubscript{R}[\degree/f]}
\\
\midrule
\multicolumn{1}{c|}{0000} & 1.32 & \textbf{0.04} & \multicolumn{1}{c|}{0.06} & 1.35 & \textbf{0.04} & \multicolumn{1}{c||}{0.06} & - & 0.05 & \multicolumn{1}{c|}{\textbf{0.05}} & \textbf{1.29} & \textbf{0.04} & \multicolumn{1}{c}{0.06} 
\\
\multicolumn{1}{c|}{0001} & \textbf{1.95} & \textbf{0.05} & \multicolumn{1}{c|}{0.04} & 2.42 & \textbf{0.05} & \multicolumn{1}{c||}{0.04} & - & 0.12 & \multicolumn{1}{c|}{0.04} & 2.31 & \textbf{0.05} & \multicolumn{1}{c}{0.04} 
\\
\multicolumn{1}{c|}{0002} & 0.95 & 0.04 & \multicolumn{1}{c|}{0.03} & 1.04 & 0.04 & \multicolumn{1}{c||}{0.03} & - & 0.04 & \multicolumn{1}{c|}{\textbf{0.02}} & \textbf{0.91} & 0.04 & \multicolumn{1}{c}{\textbf{0.02}} 
\\
\multicolumn{1}{c|}{0003} & 0.74 & 0.07 & \multicolumn{1}{c|}{0.04} & 0.78 & 0.07 & \multicolumn{1}{c||}{0.04} & - & 0.09 & \multicolumn{1}{c|}{0.04} & \textbf{0.69} & \textbf{0.06} & \multicolumn{1}{c}{0.04} 
\\
\multicolumn{1}{c|}{0004} & 1.44 & \textbf{0.07} & \multicolumn{1}{c|}{0.06} & 1.52 & \textbf{0.07} & \multicolumn{1}{c||}{0.06} & - & 0.11 & \multicolumn{1}{c|}{\textbf{0.05}} & \textbf{1.42} & \textbf{0.07} & \multicolumn{1}{c}{0.06} 
\\
\multicolumn{1}{c|}{0005} & 1.23 & \textbf{0.06} & \multicolumn{1}{c|}{0.03} & \textbf{1.22} & \textbf{0.06} & \multicolumn{1}{c||}{0.03} & - & 0.10 & \multicolumn{1}{c|}{\textbf{0.02}} & 1.34 & \textbf{0.06} & \multicolumn{1}{c}{0.03} 
\\
\multicolumn{1}{c|}{0006} & 0.19 & 0.02 & \multicolumn{1}{c|}{\textbf{0.04}} & 0.19 & 0.02 & \multicolumn{1}{c||}{\textbf{0.04}} & - & 0.02 & \multicolumn{1}{c|}{0.05} & 0.19 & 0.02 & \multicolumn{1}{c}{\textbf{0.04}} 
\\
\multicolumn{1}{c|}{0007} & \textbf{2.47} & 0.05 & \multicolumn{1}{c|}{0.07} & 2.69 & 0.05 & \multicolumn{1}{c||}{0.07} & - & - & \multicolumn{1}{c|}{-} & 3.10 & 0.05 & \multicolumn{1}{c}{0.07} 
\\
\multicolumn{1}{c|}{0008} & 1.40 & \textbf{0.08} & \multicolumn{1}{c|}{0.04} & \textbf{1.29} & \textbf{0.08} & \multicolumn{1}{c||}{0.04} & - & - & \multicolumn{1}{c|}{-} & 1.68 & 0.10 & \multicolumn{1}{c}{0.04} 
\\
\multicolumn{1}{c|}{0009} & 4.00 & 0.06 & \multicolumn{1}{c|}{\textbf{0.05}} & \textbf{3.55} & 0.06 & \multicolumn{1}{c||}{\textbf{0.05}} & - & - & \multicolumn{1}{c|}{-} & 5.02 & 0.06 & \multicolumn{1}{c}{0.06} 
\\
\multicolumn{1}{c|}{0010} & 1.68 & 0.07 & \multicolumn{1}{c|}{0.04} & 1.84 & 0.07 & \multicolumn{1}{c||}{0.04} & - & - & \multicolumn{1}{c|}{-} & \textbf{1.30} & 0.07 & \multicolumn{1}{c}{\textbf{0.03}} 
\\
\multicolumn{1}{c|}{0011} & \textbf{0.97} & 0.04 & \multicolumn{1}{c|}{0.03} & 1.05 & 0.04 & \multicolumn{1}{c||}{0.03} & - & - & \multicolumn{1}{c|}{-} & 1.03 & 0.04 & \multicolumn{1}{c}{0.03} 
% \\
% \multicolumn{1}{c|}{0012} & 0.01 & 0.00 & \multicolumn{1}{c|}{0.01} & 0.01 & 0.00 & \multicolumn{1}{c||}{0.01} & - & - & \multicolumn{1}{c|}{-} & 0.01 & 0.00 & \multicolumn{1}{c}{0.01} 
\\
\multicolumn{1}{c|}{0013} & 1.18 & 0.04 & \multicolumn{1}{c|}{0.05} & 1.18 & 0.04 & \multicolumn{1}{c||}{0.05} & - & - & \multicolumn{1}{c|}{-} & \textbf{1.10} & 0.04 & \multicolumn{1}{c}{\textbf{0.04}} 
\\
\multicolumn{1}{c|}{0014} & 0.13 & 0.03 & \multicolumn{1}{c|}{0.08} & 0.13 & 0.03 & \multicolumn{1}{c||}{0.08} & - & - & \multicolumn{1}{c|}{-} & \textbf{0.12} & 0.03 & \multicolumn{1}{c}{0.08} 
% \\
% \multicolumn{1}{c|}{0015} & 0.23 & 0.02 & \multicolumn{1}{c|}{0.02} & 0.22 & 0.02 & \multicolumn{1}{c||}{0.02} & - & - & \multicolumn{1}{c|}{-} & 0.22 & 0.02 & \multicolumn{1}{c}{0.02} 
% \\
% \multicolumn{1}{c|}{0016} & 0.02 & 0.00 & \multicolumn{1}{c|}{0.01} & 0.02 & 0.00 & \multicolumn{1}{c||}{0.01} & - & - & \multicolumn{1}{c|}{-} & 0.02 & 0.00 & \multicolumn{1}{c}{0.01} 
% \\
% \multicolumn{1}{c|}{0017} & 0.04 & 0.00 & \multicolumn{1}{c|}{0.01} & 0.04 & 0.00 & \multicolumn{1}{c||}{0.01} & - & - & \multicolumn{1}{c|}{-} & 0.04 & 0.00 & \multicolumn{1}{c}{0.01} 
\\
\multicolumn{1}{c|}{0018} & \textbf{0.89} & \textbf{0.05} & \multicolumn{1}{c|}{0.03} & 1.00 & \textbf{0.05} & \multicolumn{1}{c||}{0.03} & - & 0.07 & \multicolumn{1}{c|}{\textbf{0.02}} & 1.09 & \textbf{0.05} & \multicolumn{1}{c}{\textbf{0.02}} 
\\
\multicolumn{1}{c|}{0019} & 2.31 & 0.05 & \multicolumn{1}{c|}{0.03} & 2.35 & 0.05 & \multicolumn{1}{c||}{0.03} & - & - & \multicolumn{1}{c|}{-} & \textbf{2.25} & 0.05 & \multicolumn{1}{c}{0.03} 
\\
\multicolumn{1}{c|}{0020} & 16.80 & 0.11 & \multicolumn{1}{c|}{0.07} & \textbf{1.10} & \textbf{0.05} & \multicolumn{1}{c||}{0.04} & - & 0.16 & \multicolumn{1}{c|}{\textbf{0.03}} & 1.36 & 0.07 & \multicolumn{1}{c}{0.04} 
\\
\midrule
\multicolumn{1}{c|}{mean} & 2.33 & 0.055 & \multicolumn{1}{c|}{0.046} & \textbf{1.45} & \textbf{0.051} & \multicolumn{1}{c||}{0.045} & - & 0.084 & \multicolumn{1}{c|}{\textbf{0.036}} & 1.54 & 0.053 & \multicolumn{1}{c}{0.043}
\\
% \midrule
\bottomrule
\end{tabularx}
\caption[Egomotion comparison on the KITTI tracking dataset.]{\label{tab: egomotion_KITTI_tracking} Egomotion comparison on the KITTI \underline{tracking} dataset. 
Results of sequences without egomotion are not shown.}
\end{table*}

\subsection{Visual Odometry}
\label{subsec: vo_experiments}

For the visual odometry experiments we have chosen the KITTI tracking (Table~\ref{tab: egomotion_KITTI_tracking}) and raw (Table~\ref{tab: egomotion_KITTI_raw}) datasets~\cite{geiger2013vision}. 
They contain gray-scale and RGB stereo sequences of urban and road scenes recorded from a car perspective with circulating vehicles and pedestrians, as well as its GPS data. 

\begin{table*}[t]
\scriptsize
\begin{tabularx}{\linewidth}{@{}l*{16}{C}}
\toprule
\multicolumn{1}{c|}{seq} & \multicolumn{3}{c|}{ORB-SLAM2} & \multicolumn{3}{c||}{DynaSLAM} & \multicolumn{3}{c|}{ClusterSLAM} & \multicolumn{3}{c|}{ClusterVO} & \multicolumn{3}{c}{Ours}
\\
\multicolumn{1}{c|}{} & ATE~[m] & RPE\textsubscript{t}[m] & \multicolumn{1}{c|}{RPE\textsubscript{R}[rd]} & ATE~[m] & RPE\textsubscript{t}[m] & \multicolumn{1}{c||}{RPE\textsubscript{R}[rd]} & ATE~[m] & RPE\textsubscript{t}[m] & \multicolumn{1}{c|}{RPE\textsubscript{R}[rd]} & ATE~[m] & RPE\textsubscript{t}[m] & \multicolumn{1}{c|}{RPE\textsubscript{R}[rd]} & ATE~[m] & RPE\textsubscript{t}[m] & \multicolumn{1}{c}{RPE\textsubscript{R}[rd]}
\\
\midrule
\multicolumn{1}{c|}{0926-0009} & 0.83 & 1.85 & \multicolumn{1}{c|}{\textbf{0.01}} & 0.81 & \textbf{1.80} & \multicolumn{1}{c||}{\textbf{0.01}} & 0.92 & 2.34 & \multicolumn{1}{c|}{0.03} & \textbf{0.79} & 2.98 & \multicolumn{1}{c|}{0.03} & 0.85 & 1.87 & \multicolumn{1}{c}{\textbf{0.01}}
\\
\multicolumn{1}{c|}{0926-0013} & 0.32 & 1.04 & \multicolumn{1}{c|}{0.01} & 0.30 & 0.99 & \multicolumn{1}{c||}{0.01} & 2.12  & 5.50 & \multicolumn{1}{c|}{0.07} & \textbf{0.26} & 1.16 & \multicolumn{1}{c|}{0.01} & 0.29 & \textbf{0.93} & \multicolumn{1}{c}{\textbf{0.00}}
\\
\multicolumn{1}{c|}{0926-0014} & 0.50 & 1.22 & \multicolumn{1}{c|}{\textbf{0.01}} & 0.60 & 1.62 & \multicolumn{1}{c||}{\textbf{0.01}} & 0.81 & 2.24 & \multicolumn{1}{c|}{0.03} & \textbf{0.48} & \textbf{1.04} & \multicolumn{1}{c|}{\textbf{0.01}} & \textbf{0.48} & 1.35 & \multicolumn{1}{c}{\textbf{0.01}}
\\
\multicolumn{1}{c|}{0926-0051} & \textbf{0.38} & 1.16 & \multicolumn{1}{c|}{\textbf{0.00}} & 0.46 & 1.17 & \multicolumn{1}{c||}{\textbf{0.00}} & 1.19 & 1.44 & \multicolumn{1}{c|}{0.03} & 0.81 & 2.74 & \multicolumn{1}{c|}{0.02} & 0.44 & \textbf{1.14} & \multicolumn{1}{c}{\textbf{0.00}}
\\
\multicolumn{1}{c|}{0926-0101} & \textbf{2.97} & 13.63 & \multicolumn{1}{c|}{0.03} & 3.52 & 15.14 & \multicolumn{1}{c||}{0.03} & 4.02 & \textbf{12.43} & \multicolumn{1}{c|}{\textbf{0.02}} & 3.18 & 12.78 & \multicolumn{1}{c|}{\textbf{0.02}} & 4.33 & 15.02 & \multicolumn{1}{c}{0.04}
\\
\multicolumn{1}{c|}{0929-0004} & 0.62 & 1.38 & \multicolumn{1}{c|}{\textbf{0.01}} & 0.56 & \textbf{1.36} & \multicolumn{1}{c||}{\textbf{0.01}} & 1.12 & 2.78 & \multicolumn{1}{c|}{0.02} & \textbf{0.40} & 1.77 & \multicolumn{1}{c|}{0.02} & 0.64 & 1.41 & \multicolumn{1}{c}{\textbf{0.01}}
\\
\multicolumn{1}{c|}{1003-0047} & 20.49 & 32.59 & \multicolumn{1}{c|}{0.08} & \textbf{2.87} & \textbf{5.95} & \multicolumn{1}{c||}{\textbf{0.02}} & 10.21 & 8.94 & \multicolumn{1}{c|}{0.06} & 4.79 & 6.54 & \multicolumn{1}{c|}{0.05} & 3.03 & 6.85 & \multicolumn{1}{c}{\textbf{0.02}}
\\
\midrule
\multicolumn{1}{c|}{mean} & 3.73 & 7.55 & \multicolumn{1}{c|}{0.02} & \textbf{1.30} & \textbf{4.00} & \multicolumn{1}{c||}{\textbf{0.01}} & 2.91 & 5.10 & \multicolumn{1}{c|}{0.04} & 1.53 & 4.14 & \multicolumn{1}{c|}{0.02} & 1.44 & 4.08 & \multicolumn{1}{c}{\textbf{0.01}}
\\
\bottomrule
\end{tabularx}
\caption[Egomotion comparison on the KITTI raw dataset]{\label{tab: egomotion_KITTI_raw} Egomotion comparison on the KITTI \underline{raw} dataset}
\end{table*}

Both tables detail comparisons of our system's performance against ORB-SLAM2 and our previous work DynaSLAM~\cite{bescos2018dynaslam}. 
ORB-SLAM2 is the base SLAM system on which we build DynaSLAM II, and does not specifically address dynamic objects. 
DynaSLAM adds ORB-SLAM2 the capability to detect the features belonging to dynamic objects and classes but uniquely ignores them and does not track them. 
The difference in the results of ORB-SLAM2 and DynaSLAM gives an idea of how dynamic each sequence is. 
Theoretically, if dynamic objects are representative in the scene and they are in circulation, DynaSLAM has better performance, as can be seen in sequences 0020 and 1003-0047 in Tables~\ref{tab: egomotion_KITTI_tracking} and~\ref{tab: egomotion_KITTI_raw} respectively. 
However, if dynamic objects are representative in the scene but not in motion, \textit{e.g.}, parked cars, DynaSLAM shows a larger trajectory error. 
This happens because the features belonging to the static vehicles, which are useful for pose estimation and usually lay in nearby scene regions, are not utilized. 
This can be seen for example in the sequence 0001 in Table~\ref{tab: egomotion_KITTI_tracking}. 
Besides that, DynaSLAM II achieves a performance better than both ORB-SLAM and DynaSLAM in these two types of scenarios in many of the evaluated sequences. 
On the one hand, when dynamic instances are moving, DynaSLAM~II successfully estimates the velocity of the corresponding objects and provide the BA with rich clues for camera pose estimation when the static representation is not sufficient. 
This often occurs when dynamic objects occlude nearby scene regions and thus static features only provide valuable hints for accurately estimating the camera rotation. 
On the other hand, when dynamic classes instances are static, DynaSLAM~II tracks their features estimating that their velocity is close to zero. 
Consequently, these object points act much like static points. 

Tables~\ref{tab: egomotion_KITTI_tracking} and~\ref{tab: egomotion_KITTI_raw} present our ego motion results compared to those of state-of-the-art systems that also track dynamic objects in a joint SLAM framework. 
ClusterSLAM~\cite{huang2019clusterslam} acts as a back end rather than a SLAM system and is highly dependent on the camera poses initial estimates. 
ClusterVO~\cite{huang2020clustervo} and VDO-SLAM~\cite{zhang2020vdo} are SLAM systems as ours with the multi-object tracking capability. 
The former can handle stereo and \mbox{RGB-D} data, whereas the latter only handles \mbox{RGB-D}.
The reported errors are given with different metrics so that we can directly use the values that the authors provide. 
DynaSLAM II achieves in all sequences a lower translational relative error ($RPE_t$) than that of VDO-SLAM. 
However, VDO-SLAM usually achieves a lower rotational pose error. 
Since far points are the ones that provide the richest clues for rotation estimation, we believe that this difference in accuracy does not depend on the object tracking performance and is therefore due to the underlying camera pose estimation algorithm and sensor suite. 
Regarding the performance of ClusterVO, it achieves an accuracy which is in most sequences quite similar to ours.

\subsection{Multi-Object Tracking}
\label{subsec: mot_experiments}

Once the utility of tracking dynamic objects for ego motion estimation is demonstrated, we have chosen once again the KITTI tracking dataset~\cite{geiger2013vision} to validate our multi-object tracking results . 
The trajectories and the 3D bounding boxes of the dynamic objects are provided thanks to expensive manual annotations on LIDAR 3D point clouds. 

First of all, we would like to draw the attention of the reader to Fig.~\ref{fig: teaser_image} to have a look at our qualitative results on this dataset. 
The bounding boxes of the two purple cars on the left are well estimated despite their partial view. 
This scene is also challenging because the other two front cars are far from the camera and yet are correctly tracked.

In the last decade Bernardin \textit{et~al.}~\cite{bernardin2008evaluating} introduced the CLEAR MOT metrics to allow for objective comparison of tracker characteristics, focusing on their precision in estimating object locations, their accuracy in recognizing object configurations and their ability to consistently label objects over time. 
Whereas these metrics are well established in the computer vision and robotics communities and provide valuable insights about the per-frame performance of trackers, they do not take into account the quality of the tracked object trajectories. 
We suggest that to correctly evaluate multi-object tracking within a SLAM framework, one needs to report the CLEAR MOT metric MOTP~\footnote{MOTP stands for multiple object tracking precision. 
It is the predictions precision computed with any given cost function over the number of TPs.} as well as the common trajectory error metrics. 
Most related works on SLAM and multi-object tracking only report the CLEAR MOT metric MOTP~\cite{yang2019cubeslam, huang2020clustervo, li2018stereo} and besides that, the authors of VDO-SLAM~\cite{zhang2020vdo} uniquely report the relative pose error of all objects trajectories of one sequence as a single ensemble. 
We think that, to facilitate comparison, this metric should be instead reported for individual trajectories.

Table~\ref{tab: object_tracking_ap} shows an evaluation of all object detections in the KITTI tracking dataset with the KITTI 3D object detection benchmark. 
This allows us to directly compare our multi-object tracking results to those of other state-of-the-art similar systems (Table~\ref{tab: object_tracking_ap}). 
The CNNs of Chen \textit{et~al.}~\cite{chen20173d} and specially of Li \textit{et~al.}~\cite{li2018stereo} achieve excellent results thanks to the single-view network accuracy itself and the multi-view refinement approach of the latter one, to the detriment of a generality loss. 
On the other hand, the accuracy of Barsan \textit{et~al.}~\cite{barsan2018robust} and of Huang \textit{et~al.}~\cite{huang2020clustervo} in detecting bounding boxes is remarkable, but is very sensitive to object truncation and occlusion. 
Our results show that we can handle objects truncation and occlusion with a minor loss in precision. 
However, less bounding boxes are usually discovered. 
Our intuition is that our system feature-based nature renders this step specially challenging, opposite to the work by Barsan \textit{et~al.}~\cite{barsan2018robust}, which first computes dense stereo matching.

\begin{table}[h]
    \footnotesize
    \centering
    \begin{tabularx}{\linewidth}{@{}l*{7}{C}}
    \toprule
    \multicolumn{1}{c|}{} & \multicolumn{3}{c|}{MOTP\textsubscript{BV}} & \multicolumn{3}{c}{MOTP\textsubscript{3D}}
    \\
    \multicolumn{1}{c|}{} & Easy & Moderate & \multicolumn{1}{c|}{Hard} & Easy & Moderate & Hard
    \\
    \midrule
    \multicolumn{1}{c|}{\cite{chen20173d}} & 81.34~\% & 70.70~\% & \multicolumn{1}{c|}{66.32 \%} & 80.62~\% & 70.01~\% & 65.76~\%
    \\
    \multicolumn{1}{c|}{\cite{li2018stereo}} & \textbf{88.07~\%} & \textbf{77.83~\%} & \multicolumn{1}{c|}{\textbf{72.73~\%}} & \textbf{86.57~\%} & \textbf{74.13~\%} & \textbf{68.96~\%}
    \\
    \midrule
    \multicolumn{1}{c|}{\cite{barsan2018robust}} & 71.83~\% & 47.16~\% & \multicolumn{1}{c|}{40.30~\%} & \textbf{64.51}~\% & 43.70~\% & 37.66~\%
    \\
    \multicolumn{1}{c|}{\cite{huang2020clustervo}} & \textbf{74.65~\%} & 49.65~\% & \multicolumn{1}{c|}{45.62~\%} & 55.85~\% & 38.93~\% & 33.55~\%
    \\
    \multicolumn{1}{c|}{Ours} & 64.69~\% & \textbf{58.75~\%} & \multicolumn{1}{c|}{\textbf{58.36~\%}} & 53.14~\% & \textbf{48.66~\%} & \textbf{48.57~\%}
    \\
    \bottomrule
    \end{tabularx}
    \caption[MOTP evaluation on the KITTI tracking dataset.]{MOTP evaluation on the KITTI tracking dataset. 
    The categories Easy, Moderate and Hard are based on the 2D bounding boxes height, occlusion and truncation level.}
    \label{tab: object_tracking_ap}
\end{table}

To evaluate our estimation of object trajectories, in Table~\ref{tab: object_tracking_trajectories} we have chosen the 12 longest sequences of the KITTI tracking dataset whose 2D detections are neither occluded nor truncated, and whose height is at least 40 pixels. 
These chosen objects are labeled with their ground-truth object id. 
For each of these ground truth trajectories we look for the most overlapping bounding boxes in our estimations (the overlapping has to be of at least 25~\%). 
In the case of the trajectory metrics (ATE and RPE) and the 2D MOTP, this overlapping is computed as the IoU of the 3D bounding boxes projected over the current frame. 
For the other two evaluations (BV and 3D), the overlapping is computed as the IoU of the bounding boxes in bird view and in 3D respectively. 
This evaluation gives an idea of our framework tracking performance and our bounding boxes quality. 
Regarding the true positives percentage, we can see that objects are tracked for the majority of their trajectory. 
Missing detections occur because the objects lay far from the camera and the stereo matching does not provide enough features for a rich tracking. 
It is important to notice that the accuracy of the passersby tracking is lower than that of the cars due to their non-rigid shape (sequence 0017). 
The trajectory errors of the cars are acceptable but they are far from the ego-motion estimation performance. 
Our intuition is that our algorithm feature-based nature renders the bounding box estimation specially challenging. 
A larger amount of 3D points would always provide richer clues for object tracking.

\begin{table*}[t]
\scriptsize
\begin{tabularx}{\linewidth}{@{}l*{13}{C}}
\toprule
\multicolumn{2}{c|}{sequence} & \multicolumn{1}{c|}{$0003$} & \multicolumn{1}{c|}{$0005$} & \multicolumn{1}{c|}{$0010$} & \multicolumn{2}{c|}{$0011$} & \multicolumn{2}{c|}{$0018$} & \multicolumn{2}{c|}{$0019$} & \multicolumn{3}{c}{$0020$}
\\
\multicolumn{2}{c|}{object id (class)} & \multicolumn{1}{c|}{$1$ (car)} & \multicolumn{1}{c|}{$31$ (car)} & \multicolumn{1}{c|}{$0$ (car)} & $0$ (car) & \multicolumn{1}{c|}{$35$ (car)} & $2$ (car) & \multicolumn{1}{c|}{$3$ (car)} & $63$ (car) & \multicolumn{1}{c|}{$72$ (car)} & $0$ (car) & $12$ (car) & $122$~(car)
\\
\midrule
\multicolumn{2}{c|}{ATE~[$m$]} & \multicolumn{1}{c|}{0.69} & \multicolumn{1}{c|}{0.51} & \multicolumn{1}{c|}{0.95} & 1.05 & \multicolumn{1}{c|}{1.25} & 1.10 & \multicolumn{1}{c|}{1.13} & 0.86 & \multicolumn{1}{c|}{0.99} & 0.56 & 1.18 & 0.87
\\
\multicolumn{2}{c|}{RPE\textsubscript{t}~[$m/m$]} & \multicolumn{1}{c|}{0.34} & \multicolumn{1}{c|}{0.26} & \multicolumn{1}{c|}{0.40} & 0.43 & \multicolumn{1}{c|}{0.89} & 0.30 & \multicolumn{1}{c|}{0.55} & 1.45 & \multicolumn{1}{c|}{1.12} & 0.45 & 0.40 & 0.72
\\
\multicolumn{2}{c|}{RPE\textsubscript{R}~[$\degree/m$]} & \multicolumn{1}{c|}{1.84} & \multicolumn{1}{c|}{13.50} & \multicolumn{1}{c|}{2.84} & 12.51 & \multicolumn{1}{c|}{16.64} & 9.27 & \multicolumn{1}{c|}{20.05} & 48.80 & \multicolumn{1}{c|}{3.36} & 1.30 & 6.19 & 5.75
\\
\midrule
\midrule
\multirow{2}{*}{2D} & \multicolumn{1}{c|}{TP ($\%$)} & \multicolumn{1}{c|}{50.00} & \multicolumn{1}{c|}{28.96} & \multicolumn{1}{c|}{81.63} & 72.65 & \multicolumn{1}{c|}{53.17} & 86.36 & \multicolumn{1}{c|}{53.33} & 35.26 & \multicolumn{1}{c|}{29.11} & 63.68 & 42.77 & 34.90
\\
& \multicolumn{1}{c|}{MOTP~[$\%$]} & \multicolumn{1}{c|}{71.79} & \multicolumn{1}{c|}{60.30} & \multicolumn{1}{c|}{73.51} & 74.78 & \multicolumn{1}{c|}{65.25} & 74.81 & \multicolumn{1}{c|}{70.94} & 63.50 & \multicolumn{1}{c|}{62.59} & 78.54 & 76.77 & 78.76
\\
\midrule
\multirow{2}{*}{BV} & \multicolumn{1}{c|}{TP ($\%$)} & \multicolumn{1}{c|}{39.34} & \multicolumn{1}{c|}{14.48} & \multicolumn{1}{c|}{70.41} & 61.66 & \multicolumn{1}{c|}{19.05} & 67.05 & \multicolumn{1}{c|}{21.75} & 29.48 & \multicolumn{1}{c|}{29.43} & 43.78 & 37.64 & 34.51
\\
& \multicolumn{1}{c|}{MOTP~[$\%$]} & \multicolumn{1}{c|}{56.61} & \multicolumn{1}{c|}{46.84} & \multicolumn{1}{c|}{47.60} & 50.74 & \multicolumn{1}{c|}{31.95} & 45.47 & \multicolumn{1}{c|}{41.45} & 45.69 & \multicolumn{1}{c|}{55.48} & 45.00 & 49.29 & 48.05
\\
\midrule
\multirow{2}{*}{3D} & \multicolumn{1}{c|}{TP ($\%$)} & \multicolumn{1}{c|}{38.53} & \multicolumn{1}{c|}{11.45} & \multicolumn{1}{c|}{68.37} & 52.28 & \multicolumn{1}{c|}{6.35} & 62.12 & \multicolumn{1}{c|}{16.84} & 26.48 & \multicolumn{1}{c|}{29.43} & 31.84 & 36.23 & 29.02
\\
& \multicolumn{1}{c|}{MOTP~[$\%$]} & \multicolumn{1}{c|}{48.20} & \multicolumn{1}{c|}{34.20} & \multicolumn{1}{c|}{40.28} & 47.35 & \multicolumn{1}{c|}{26.02} & 34.80 & \multicolumn{1}{c|}{35.80} & 33.89 & \multicolumn{1}{c|}{39.81} & 46.15 & 40.81 & 44.43
\\
\bottomrule
\end{tabularx}
\caption{\label{tab: object_tracking_trajectories} Objects motion comparison on the KITTI \underline{tracking} dataset.}
\end{table*}

\subsection{Timing Analysis}

To complete the evaluation of our proposal, Table~\ref{tab: TrackSLAM_time} shows the average computational time for its different building blocks. 
The timing of DynaSLAM II is highly dependent on the number of objects to be tracked. 
In sequences like KITTI tracking 0003 there are only two objects at a time as maximum and it can thus run at $12$ fps. 
However, the sequence 0020 can have up to 20 objects at a time and its performance is seen slightly compromised, but still achieves a real time performance at $\sim10$ fps. 
We do not include within these numbers the computational time of the semantic segmentation CNN since it depends on the GPU power and CNN model complexity. 
Algorithms such as YOLACT~\cite{bolya2019yolact} can run in real time and provide high-quality instance masks.

\begin{table}[t]
    \footnotesize
    \centering
    \begin{tabularx}{\linewidth}{@{}l*{6}{C}}
        \toprule
        Sequence & \multicolumn{3}{|c|}{Building block} & \multicolumn{2}{c}{Time [ms]}
        \\
        \midrule
        \multirow{3}{*}{KITTI tracking 0003} & \multicolumn{3}{|c|}{Tracking thread} & \multicolumn{2}{c}{80.10 $\pm$ 0.78} 
        \\
         & \multicolumn{3}{|c|}{Local BA} & \multicolumn{2}{c}{61.37 $\pm$ 6.70}
        \\
          & \multicolumn{3}{|c|}{Bounding Boxes BA} & \multicolumn{2}{c}{0.07 $\pm$ 0.01}
        \\
        % \midrule
        % \multirow{3}{*}{KITTI tracking 0005} & \multicolumn{3}{|c|}{Tracking thread} & \multicolumn{2}{c}{86.46 \pm 0.82} 
        % \\
        %  & \multicolumn{3}{|c|}{Local BA} & \multicolumn{2}{c}{69.90 \pm 16.87}
        % \\
        %  & \multicolumn{3}{|c|}{Bounding Boxes BA} & \multicolumn{2}{c}{0.17 \pm 0.02}
        % \\
        \midrule
        \multirow{3}{*}{KITTI tracking 0020} & \multicolumn{3}{|c|}{Tracking thread} & \multicolumn{2}{c}{94.56 $\pm$ 1.27} 
        \\
         & \multicolumn{3}{|c|}{Local BA} & \multicolumn{2}{c}{65.03 $\pm$ 17.72}
        \\
          & \multicolumn{3}{|c|}{Bounding Boxes BA} & \multicolumn{2}{c}{0.60 $\pm$ 0.05}
        \\
        \midrule
        \midrule
        \multicolumn{1}{c|}{} & \multicolumn{1}{c|}{\cite{li2018stereo}} & \multicolumn{1}{c|}{\cite{zhang2020vdo}} & \multicolumn{1}{c|}{\cite{huang2019clusterslam}} & \multicolumn{1}{c|}{\cite{huang2020clustervo}} & Ours \\
        \midrule
        \multicolumn{1}{c|}{fps} & \multicolumn{1}{c|}{5.8} & \multicolumn{1}{c|}{5 - 8} & \multicolumn{1}{c|}{7} & \multicolumn{1}{c|}{8} & \textbf{10 - 12} \\
        \bottomrule
    \end{tabularx}
    \caption[DynaSLAM II average computational time.]{DynaSLAM II average computational time.}
    \label{tab: TrackSLAM_time}
\end{table}

Finally, the last rows of Table~\ref{tab: TrackSLAM_time} collect the average timing results for systems that jointly perform SLAM and multi-object tracking in the KITTI dataset. 
DynaSLAM II is the only system that can provide at present a real-time solution.

\section{Conclusions and Future Work}
\label{sec: conclusions}

We have proposed an object level SLAM system with novel measurement functions between cameras, objects and 3D map points. 
This allows us to track dynamic objects and tightly optimize the trajectories of self and surroundings to let both estimations be mutually beneficial.
We decouple the problem of object tracking from that of bounding boxes estimation and, differently from other works, we do not make any assumptions about the objects motion, pose or model. 
Our experiments show that DynaSLAM~II achieves a state-of-the-art accuracy at real time performance, which renders our framework suitable for a large number of applications. 

The feature-based core of our system limits its ability to discover accurate 3D bounding boxes, and also to track objects with low texture. 
Fully exploiting the dense visual information would certainly push these limits forward. 
We would also like to explore the --even more-- challenging task of multi-object tracking and SLAM with only a monocular camera. 
This is an interesting direction since dynamic object tracking can provide rich clues about the scale of the map.

% \input{chapters/extra_stuff}

% \input{chapters/appendix}


\bibliographystyle{ieeetr}
\bibliography{references}

\begin{thebibliography}{10}

\bibitem{mur2017orb}
R.~Mur-Artal and J.~D. Tard{\'o}s, ``{ORB-SLAM2}: An open-source {SLAM} system
  for monocular, stereo, and {RGB-D} cameras,'' {\em IEEE T-RO}, 2017.

\bibitem{engel2017direct}
J.~Engel, V.~Koltun, and D.~Cremers, ``Direct sparse odometry,'' {\em
  Transactions on pattern analysis and machine intelligence}, 2017.

\bibitem{forster2014svo}
C.~Forster, M.~Pizzoli, and D.~Scaramuzza, ``{SVO: Fast semi-direct monocular
  visual odometry},'' in {\em IEEE ICRA}, pp.~15--22, 2014.

\bibitem{bescos2018dynaslam}
B.~Bescos, J.~M. F{\'a}cil, J.~Civera, and J.~Neira, ``{DynaSLAM: Tracking,
  mapping, and inpainting in dynamic scenes},'' {\em IEEE RA-L}, 2018.

\bibitem{sun2017improving}
Y.~Sun, M.~Liu, and M.~Q.-H. Meng, ``Improving {RGB-D} {SLAM} in dynamic
  environments: A motion removal approach,'' {\em RAS}, 2017.

\bibitem{li2017rgb}
S.~Li and D.~Lee, ``{RGB-D SLAM in dynamic environments using static point
  weighting},'' {\em IEEE Robotics and Automation Letters}, 2017.

\bibitem{xiao2019dynamic}
L.~Xiao, J.~Wang, X.~Qiu, Z.~Rong, and X.~Zou, ``{Dynamic-SLAM: Semantic
  monocular visual localization and mapping based on deep learning in dynamic
  environment},'' {\em RAS}, vol.~117, 2019.

\bibitem{bescos2019empty}
B.~Bescos, J.~Neira, R.~Siegwart, and C.~Cadena, ``Empty cities: Image
  inpainting for a dynamic-object-invariant space,'' in {\em IEEE ICRA}, 2019.

\bibitem{bevsic2020dynamic}
B.~Be{\v{s}}i{\'c} and A.~Valada, ``{Dynamic Object Removal and Spatio-Temporal
  RGB-D Inpainting via Geometry-Aware Adversarial Learning},'' {\em
  arXiv:2008.05058}, 2020.

\bibitem{li2018stereo}
P.~Li, T.~Qin, {\em et~al.}, ``{Stereo vision-based semantic 3D object and
  ego-motion tracking for autonomous driving},'' in {\em IEEE ECCV}, 2018.

\bibitem{yang2019cubeslam}
S.~Yang and S.~Scherer, ``{CubeSLAM: Monocular 3-D object SLAM},'' {\em IEEE
  Transactions on Robotics}, vol.~35, no.~4, pp.~925--938, 2019.

\bibitem{huang2020clustervo}
J.~Huang, S.~Yang, T.-J. Mu, and S.-M. Hu, ``{ClusterVO: Clustering Moving
  Instances and Estimating Visual Odometry for Self and Surroundings},'' in
  {\em IEEE CVPR}, pp.~2168--2177, 2020.

\bibitem{zhang2020vdo}
J.~Zhang, M.~Henein, R.~Mahony, and V.~Ila, ``{VDO-SLAM: A Visual Dynamic
  Object-aware SLAM System},'' {\em arXiv:2005.11052}, 2020.

\bibitem{wang2003online}
C.-C. Wang, C.~Thorpe, and S.~Thrun, ``Online simultaneous localization and
  mapping with detection and tracking of moving objects: Theory and results
  from a ground vehicle in crowded urban areas,'' in {\em IEEE International
  Conference on Robotics and Automation}, 2003.

\bibitem{wangsiripitak2009avoiding}
S.~Wangsiripitak and D.~W. Murray, ``{Avoiding moving outliers in visual SLAM
  by tracking moving objects},'' in {\em ICRA}, IEEE, 2009.

\bibitem{rogers2010slam}
J.~G. Rogers, A.~J. Trevor, C.~Nieto-Granda, and H.~I. Christensen, ``{SLAM
  with expectation maximization for moveable object tracking},'' in {\em IEEE
  International Conf. on Intelligent Robots and Systems}, 2010.

\bibitem{barsan2018robust}
I.~A. B{\^a}rsan, P.~Liu, M.~Pollefeys, and A.~Geiger, ``Robust dense mapping
  for large-scale dynamic environments,'' in {\em IEEE ICRA}, 2018.

\bibitem{rosinol20203d}
A.~Rosinol, A.~Gupta, M.~Abate, J.~Shi, and L.~Carlone, ``{3D Dynamic Scene
  Graphs: Actionable Spatial Perception with Places, Objects, and Humans},''
  {\em arXiv:2002.06289}, 2020.

\bibitem{geiger2013vision}
A.~Geiger, P.~Lenz, C.~Stiller, and R.~Urtasun, ``{Vision meets robotics: The
  kitti dataset},'' {\em IJRR}, vol.~32, no.~11, pp.~1231--1237, 2013.

\bibitem{wang2007simultaneous}
C.-C. Wang, C.~Thorpe, S.~Thrun, M.~Hebert, and H.~Durrant-Whyte,
  ``Simultaneous localization, mapping and moving object tracking,'' {\em The
  International Journal of Robotics Research}, vol.~26, no.~9, 2007.

\bibitem{runz2017co}
M.~R{\"u}nz and L.~Agapito, ``Co-fusion: Real-time segmentation, tracking and
  fusion of multiple objects,'' in {\em IEEE ICRA}, pp.~4471--4478, 2017.

\bibitem{runz2018maskfusion}
M.~Runz, M.~Buffier, and L.~Agapito, ``{Maskfusion: Real-time recognition,
  tracking and reconstruction of multiple moving objects},'' in {\em IEEE
  International Symposium on Mixed and Augmented Reality}, 2018.

\bibitem{xu2019mid}
B.~Xu, W.~Li, D.~Tzoumanikas, M.~Bloesch, A.~Davison, and S.~Leutenegger,
  ``{MID-fusion: Octree-based object-level multi-instance dynamic SLAM},'' in
  {\em IEEE ICRA}, pp.~5231--5237, 2019.

\bibitem{henein2018exploiting}
M.~Henein, G.~Kennedy, R.~Mahony, and V.~Ila, ``{Exploiting rigid body motion
  for SLAM in dynamic environments},'' {\em IEEE ICRA}, 2018.

\bibitem{huang2019clusterslam}
J.~Huang, S.~Yang, Z.~Zhao, Y.-K. Lai, and S.-M. Hu, ``{ClusterSLAM: A SLAM
  Backend for Simultaneous Rigid Body Clustering and Motion Estimation},'' in
  {\em IEEE ICCV}, pp.~5875--5884, 2019.

\bibitem{rublee2011orb}
E.~Rublee, V.~Rabaud, K.~Konolige, and G.~Bradski, ``{ORB: An efficient
  alternative to SIFT or SURF},'' in {\em ICCV}, IEEE, 2011.

\bibitem{bernardin2008evaluating}
K.~Bernardin and R.~Stiefelhagen, ``{Evaluating multiple object tracking
  performance: the CLEAR MOT metrics},'' {\em EURASIP JIVP}, 2008.

\bibitem{chen20173d}
X.~Chen, K.~Kundu, Y.~Zhu, H.~Ma, S.~Fidler, and R.~Urtasun, ``{3D object
  proposals using stereo imagery for accurate object class detection},'' {\em
  IEEE TPAMI}, vol.~40, no.~5, pp.~1259--1272, 2017.

\bibitem{bolya2019yolact}
D.~Bolya, C.~Zhou, F.~Xiao, and Y.~J. Lee, ``Yolact: Real-time instance
  segmentation,'' in {\em ICCV}, pp.~9157--9166, IEEE, 2019.

\end{thebibliography}

\end{document}